\pdfoutput=1

\documentclass[11pt]{article}

\usepackage{acl}

\usepackage{times}
\usepackage{latexsym}

\usepackage[T1]{fontenc}

\usepackage[utf8]{inputenc}

\usepackage{microtype}

\title{Slot Filling for Biomedical Information Extraction}

\date{} 					

\author{ Yannis Papanikolaou, Marlene Staib, Justin Grace and Francine Bennett\\
Healx\\
	Cambridge\\
	UK \\
	\texttt{(\small{yannis.papanikolaou, marlene.staib, justin.grace, francine.bennett)@healx.io}}
}

\begin{document}
\maketitle

\begin{abstract}
Information Extraction (IE) from text refers to the task of extracting structured knowledge from unstructured text. The task typically consists of a series of sub-tasks such as Named Entity Recognition and Relation Extraction.
Sourcing entity and relation type specific training data is a major bottleneck in domains with limited resources such as biomedicine. In this work we present a slot filling approach to the task of biomedical IE, effectively replacing the need for entity and relation-specific training data, allowing us to deal with zero-shot settings. We follow the recently proposed paradigm of coupling a Tranformer-based bi-encoder, Dense Passage Retrieval, with a Transformer-based reading comprehension model to extract relations from biomedical text. We assemble a biomedical slot filling dataset for both retrieval and reading comprehension and conduct a series of experiments demonstrating that our approach outperforms a number of simpler baselines. We also evaluate our approach end-to-end for standard as well as zero-shot settings. Our work provides a fresh perspective on how to solve biomedical IE tasks, in the absence of relevant training data. Our code, models and datasets are available at \url{https://github.com/ypapanik/biomedical-slot-filling}.
\end{abstract}

\section{Introduction}

In Information Extraction (IE) we are interested in extracting structured knowledge from unstructured text. This structured knowledge takes most usually the form of directed binary relations between entities, in other words triples of the form \textit{head - relation - tail}, which can then be used to populate a Knowledge Base or a Knowledge Graph with factual information.

The standard approach to perform IE relies on a cascade of Natural Language Processing (NLP) models. First, Named Entity Recognition (NER) is employed to find and extract entities of interest, subsequently Entity Linking (EL) to link the extracted entities to Knowledge Base identifiers and finally Relation Extraction (RE) to identify existing relations between entities. 

These individual sub-tasks tasks have attracted a great deal of attention in recent years with methods and datasets fuelling further research \cite{verga2018simultaneously, zeng2014relation, zeng2015distant, lin2016neural}. IE is largely regarded as a main facilitator of structured data reasoning, such as Knowledge Base Completion.

\subsection{Standard Information Extraction vs Slot Filling}

A major bottleneck in the above approach is that all modules (NER, EL, RE) need training data specific to the entity or relation types that we are interested in extracting. For instance, a NER model recognizing diseases needs training data annotated with the entity type $disease$ and so forth. The biomedical domain is particularly affected by these limitations, given the vast variety of entity and relation types which are commonly of interest. Additionally, sourcing training data for each sub-task and type is expensive and challenging, requiring subject matter experts. For reference, the UMLS ontology contains 125 semantic (entity) types and 54 relation types.

An alternative approach to standard IE is slot filling. The way IE is conceptualized in slot filling is highly reminiscent of open domain question answering (QA): for a given head-relation query the retriever returns a set of relevant passages, which are then fed to a reader model that then extracts a matching tail entity, the answer. By following such an approach, we can deal with zero-shot settings since, unlike standard IE, we are not seeking to recognize specific entity types or extract specific relation types, but rather do machine reading comprehension, that is, extract answers in response to queries. Importantly, this approach extends to relation types that were unseen during training, effectively reducing the need for re-training and re-deployment of a model deployed into production.

Furthermore, standard IE requires processing of every single sentence of the given corpus through its different modules (NER, EL, RE). In contrast, the computational cost of slot filling is much smaller as it performs retrieval and reading comprehension on far fewer queries to extract relations. As an example, Hetionet \cite{himmelstein2017systematic} contains around 2.25M relations, but they can be formulated in around 46k distinct queries, of the form \textit{head-relation}\footnote{In other words, if we were trying to build a KB from biomedical text that would contain these 2.25M relations, we would require to perform around 46k queries on our index to retrieve relevant documents.}.

As a final point we summarize below how the two approaches would materialize in a production setting, to make their differences more apparent. We note that standard IE might involve additional tasks, such as coreference resolution (which we do not describe here for simplicity):

Standard IE:
\begin{itemize}
    \item For each sentence, recognise entities with NER model.
    \item For each recognised entity, link to an entity identifier from a Knowledge Base, discarding entries that cannot be linked.
    \item For each sentence that contains more than one recognized entity, extract relations between the entities with a RE model.
    \item Aggregate relations per sentence, resolving potential conflicts.
\end{itemize}
Slot filling:
\begin{itemize}
 \item For each entity in the Knowledge Base and each possible relation type, consider all possible head - relation pairs and construct the relevant queries, in a form \textit{head - relation}\footnote{With this formulation a head and a tail can be used interchangeably, by just changing the relation type, e.g. a \emph{drug-treats-disease relation} can also be cast to \emph{disease-is treated by-drug} without additional training data.}.
 \item For each query, retrieve the top k relevant documents with a retriever model.
 \item For each query-retrieved document pair, perform reading comprehension, extracting zero, one or multiple answers, i.e., relation tails.
 \item For each answer, link to an entity identifier from a Knowledge Base, discarding entries that cannot be linked.
\end{itemize}

\subsection{Slot Filling: General vs Biomedical Domain}
\label{sec:gen_vs_biomedical_domain}

Although similar in most aspects, slot filling in the general domain against slot filling in the biomedical and more broadly the scientific domain differ in a few key ways. The first lies in the link between relations and entities. In the general domain, a specific relation type will often imply a specific entity type as well, whereas this rarely holds in biomedical literature. Consider for example a relation \textit{child-of} in the general domain, where we expect both head and tail of the relation to be entities of type \textit{person}, as opposed to a relation \textit{(up)regulate} in biomedicine where the head might be \textit{gene} or \textit{drug} equivalently. These nuances in the language used render the task of slot filling more challenging in biomedicine.

Another, perhaps more critical aspect relates to retrieval and more specifically how we build and evaluate on a retrieval dataset. In the general domain, a slot filling query, or more broadly a question within the QA framework, will most often have a unique answer\footnote{We are implicitly referring only to factoid queries here which is the case for most open domain QA datasets; queries of list type would have multiple answers in any case.}, whereas this rarely holds when mining the biomedical literature. For instance, consider the examples illustrated in Table \ref{tbl:general_vs_biomedical_domain_sf} coming from two well established general domain benchmarks, Natural Questions \cite{kwiatkowski2019natural} and zsRE \cite{levy2017zero} against two datasets from the biomedical domain, BioASQ \cite{tsatsaronis2015overview} and our slot filling dataset (BioSF).

\begin{table*}
\centering
\begin{tabular}{ccc}
Dataset&Query&Answer(s)\\
\hline 
\hline
NQ&when is the next deadpool movie being released&May 18, 2018\\
NQ&what was the first capital city of australia&Melbourne\\
zsRE&Elmer George [SEP] spouse&Mari Hulman George\\
zsRE&Boone River [SEP] mouth of the watercourse&Des Moines River\\
\hline
BioASQ&What are the main indications of lacosamide?&'epilepsy', 'analgesic'\\
BioASQ&Which metabolite activates AtxA?&'CO2', 'bicarbonate'\\
BioSF&sildenafil [SEP] regulator&'L765A', 'F786A', 'F820A'\\
BioSF&Amprenavir [SEP] interacts with&'rifabutin', 'ritonavir'\\
\hline
\hline
\end{tabular}
\caption{\label{tbl:general_vs_biomedical_domain_sf} Examples of queries for general domain benchmarks (NQ, zsRE) vs biomedical domain benchmarks (BioASQ, BioSF). Queries in the biomedical domain usually involve multiple valid answers, as opposed to the general domain. }
\end{table*}

This difference has a number of implications both for training and evaluation. With respect to training, one of the major successes of neural-based retrieval methods has been attributed to being able to present the model with hard negatives, i.e., examples were a previous version of the retriever (or a simpler statistical retriever) have failed. When, for example, we have a query-answer pair that mentions that Barack Obama's wife is Michelle Obama, and the model returns a passage that does not include the string "Michelle Obama", we can relatively safely consider this a false positive and use that passage as a hard negative. This helps the algorithm correct mistakes and improve. In biomedicine on the other hand, if we have an example stating that sildenafil regulates a mutation L765A, we cannot be sure that all alternative strings extracted by the model are true negatives, as there may be other valid answers that we cannot validate due to our Knowledge Base being incomplete. This compromises our ability to build gold standard training data and we are presented with a situation similar to the one encountered in distant supervision, where unlabeled examples are considered as negatives but might be positives in some cases. Practically, this leads to a noisy training set which may reduce model accuracy.

During evaluation of a biomedical retriever, we encounter the same problem, in the sense that we might obtain misleading low performance since unknown correct passages might rank higher than the known correct ones. This leads to an imperfect, i.e., "silver" quality, evaluation regime making it hard to compare approaches and models.

In this work we aim to address the challenges mentioned in the two previous subsections. Specifically, 
\begin{itemize}
    \item We provide a short review of the relevant work in Section \ref{sec:related_work}.
    \item We contribute a novel formulation of biomedical IE as a slot filling task, to address few-shot or zero-shot settings in Section \ref{sec:biomedical_slot_filling}.
    \item We release a new benchmark for biomedical slot filling, dubbed \textit{BioSF} which we describe in Section \ref{sec:biosf}.
    \item We train a biomedical dense passage retriever along with a biomedical reading comprehension model for slot filling, using BioSF. We provide the models publicly. 
    \item We present an evaluation of our approach over several baselines on BioSF, which we are able to outperform by a large margin, in Section \ref{sec:experiments}.
\end{itemize}

\section{Related Work}
\label{sec:related_work}

Recent years have witnessed a series of significant advances in the field of QA, primarily owing to the Transformer architecture \cite{vaswani2017attention} and the BERT self-supervised pre-training paradigm \cite{DBLP:conf/naacl/DevlinCLT19}. These advances, both in terms of methods \cite{chen2017reading, lin2019bert, guu2020realm, lewis2020retrieval} and datasets \cite{kwiatkowski2019natural, yang2018hotpotqa}, motivated researchers to formulate a series of different NLP tasks as open domain QA, including entity linking or relation extraction \cite{levy2017zero, petroni2021kilt}. In this work we follow this paradigm by formulating biomedical IE as a slot-filling task.

In open domain QA, given a query, a retrieval module first retrieves relevant documents from the knowledge source (such as Wikipedia). A reading comprehension module is then used to extract a span from the relevant documents, the answer. The retrieval step was, up to very recently, dominated by statistical-based approaches, namely BM25 or tf-idf \cite{chen2017reading}. ORQA \cite{lee2019latent} and REALM \cite{guu2020realm} have been the first neural based methods to clearly outperform statistical based retrieval, although they required expensive language model pre-training. Dense Passage Retrieval (DPR) \cite{karpukhin2020dense} has improved upon these methods by employing BERT-based encoders, one for the queries and one for passages. These are jointly optimized during training to classify passages as relevant versus irrelevant. This approach has proved superior to other neural based approaches and has quickly become the preferred method for open domain QA in subsequent work \cite{lewis2020retrieval, izacard2021leveraging, maillard2021multi}. 

Among the subsequent works, Retrieval Augmented Generation \cite{lewis2020retrieval} employs an architecture based on DPR and BART \cite{lewis2020bart} that is optimized end to end during finetuning, to retrieve relevant documents and generate answers to queries. Fusion-in-decoder \cite{izacard2021leveraging} employs DPR or BM25 as retrievers coupled with a T5 language model, to generate answers by attending at multiple passages simultaneously. For simplicity, we are not considering these approaches in this work, leaving their implementation for the biomedical domain for future work.

In an effort to fuel further research on this field, \citet{petroni2021kilt} introduced KILT, a new benchmark of knowledge intensive tasks, which contains among others two slot filling datasets, \textit{zero-shot RE} which was first presented in \cite{levy2017zero} and \textit{T-REx} introduced by \citet{elsahar-etal-2018-rex}. In building our biomedical slot filling dataset we largely follow the conventions and format of KILT, with the intention to ease experimentation.

Finally, \citet{glass-etal-2021-robust} have presented a RAG model specifically finetuned for slot filling on the above datasets, showing significant improvement over the generic alternatives, which were finetuned on Natural Questions (NQ).

\section{Biomedical Slot Filling}
\label{sec:biomedical_slot_filling}

Formally, let us first define the task of IE. We assume a knowledge source $K$, consisting of passages $p_i$. Furthermore, we assume there exists a Knowledge Base that contains a number of entities $e_i$. Our goal is to extract from $K$ all possible triples of the form $e_a - r_i - e_b$ where $r_i \in R$ and $R$ is the set of possible relation types. For each $e_i$ we assume that it has a specific entity type $e_t$ and that each $e_t$ can be involved in a specific subset of $R$.

Slot filling further formulates the above task as follows: we first employ a retrieval model $M_r$ that encodes all passages $p_i$ from $K$. The encoded passages are indexed to allow fast retrieval. At inference, for each $e_i$ of type $e_t$, we consider all possible relations from R and construct the relevant queries $q_i:  e_i - r_i$. Each query is then encoded and the resulting vector is used to query the index, returning the n most similar $p_i$ in terms of the maximum inner product:
\begin{equation}
    sim(q_i, p_i) = E_Q(q_i)^TE_P(p_i)
\end{equation}
where $E_Q$ is the query encoder and $E_P$ is the passage encoder. Subsequently a reader model $M_{qa}$ takes as input the above query and each of the retrieved passages and extracts zero, one or more spans, i.e., answers. Valid answers are considered as those representing an entity $e_i$. 

Here, we adopt as $M_r$ a neural, dense bi-encoder, namely DPR, which uses a different encoder for passages and queries, but any type of retriever can be used such as BM25, where $E_Q=E_P$. We initialize DPR's encoders with the ones presented in \cite{karpukhin2020dense} which were finetuned on the NQ benchmark. We subsequently train DPR on the dataset presented in Section \ref{sec:biosf}, with the following loss function:
\begin{equation}
    L(q_i, p_i^+, p_i^-) = -log\frac{e^{sim(q_i, p_i^+)}}{e^{sim(q_i, p_i^+)}+e^{sim(q_i, p_i^-)}}
\end{equation}
Unlike \cite{karpukhin2020dense}, we assume that each training instance is a $(q_i, p_i^+, p_i^-)$ tuple where $p_i^+$ is a positive, i.e., relevant passage and $p_i^-$ is a negative passage.

Regarding the reader comprehension model $M_{qa}$, we employ a pretrained BioBERT \cite{lee2019biobert} model and finetune it on the dataset of Section \ref{sec:biosf}. To finetune we follow the standard approach for question answering with BERT where the input is the concatenated query and passage separated by special token \textit{[SEP]} and the outputs are the start and end token positions within the passage. The training objective is the sum of the log-likelihoods of the correct start and end positions. For more details we refer the interested reader to \cite{DBLP:conf/naacl/DevlinCLT19}.

\begin{table*}
\centering
\begin{tabular}{cccc}
\hline 
Dataset&relation&relation types&\# instances\\ 
\hline
\hline
BioCreative V CDR \cite{li2016biocreative}&compound-disease&1&15,796\\
BioCreative VI ChemProt \cite{krallinger2017overview}&compound-protein&9&15,568\\
DDIExtraction 2013 \cite{segura2013semeval}&drug-drug&1&32,018\\
\hline
\hline
\end{tabular}
\caption{\label{tbl:bio_dataset_for_biosf} Public datasets used to build our biomedical slot filling dataset, BioSF. The relation types for the drug-drug interactions dataset have been merged into one relation dubbed \textit{interacts with}.}
\end{table*}

\section{Biomedical Slot Filling Dataset}
\label{sec:biosf}
In order to build a slot filling dataset for biomedicine, we resort to a number of publicly available biomedical NER and RE datasets, summarized in Table \ref{tbl:bio_dataset_for_biosf}. Each instance in these datasets contains the relation triple as well as the text where it was found, thus we can easily transform them in a question answering-like format for slot filling. In total, we build two datasets, one to train and evaluate the retriever and one for the reader model respectively.

Specifically for the retriever training, we use negative, i.e., null relation instances, as negatives. Additionally, we have used BM25 to add hard negatives to our dataset, exactly as \cite{karpukhin2020dense, glass-etal-2021-robust} have done previously. Although, as mentioned above, these negatives might entail some noise, similarly to when following a distant supervision approach we expect the noise to cancel out overall. Both datasets with their training, development and testing splits are released with our code. In the following, we refer to our dataset as \textit{BioSF}.

\begin{table*}
\centering
\begin{tabular}{ccccc}
\hline 
Retriever&hits@1&hits@10&hits@100&index size(Gb)\\ 
\hline
\hline
BM25&21.4&36.1&60.6&\textbf{1.1}\\
DPR-NQ \cite{karpukhin2020dense}&5.5&17.2&37.6&2.9\\
DPR-multitask \cite{maillard2021multi}&4.2&14.3&33.8&2.9\\
DPR-zsRE \cite{glass-etal-2021-robust}&7.6&19.6&37.2&2.9\\
Bio-DPR(ours)&\textbf{31.0}&\textbf{55.1}&\textbf{72.5}&2.9\\
\hline
\hline
\end{tabular}
\caption{\label{tbl:retrieval_on_1m} Evaluation results for retrieval experiments on the BioSF development set using as content one million passages from PubMed. Values in bold show statistically significant results in terms of z-test at p-value of 0.05, whereas for our model we show the average across five different DPR training runs.}
\end{table*}

\section{Experiments}
\label{sec:experiments}
In this Section we present the experiments that we conducted, followed by a discussion on their implications. We are interested in evaluating our biomedical DPR retriever, our biomedical slot filling reader and finally the end to end slot filling approach. 
\begin{table*}
\centering
\begin{tabular}{ccccc}
\hline 
Retriever&hits@1&hits@10&hits@100&index size \\ 
\hline
\hline
BM25&11.0&30.3&56.1&\textbf{29.4}\\
DPR-NQ&5.2&17.9&38.9&90.0\\
DPR-zsRE&2.3&10.2&26.4&90.0\\
Bio-DPR(ours)&\textbf{11.5}&\textbf{33.2}&\textbf{59.1}&90.0\\
\hline
\hline
\end{tabular}
\caption{\label{tbl:retrieval} Evaluation results for retrieval experiments on the BioSF development set on full PubMed. Values in bold show statistically significant results for a z-test at p-value of 0.05.}
\end{table*}

\subsection{Retrieval}
\label{sec-exp-retrieval}
First, we are interested to understand the performance of our approach against different baselines. To that end, we employ BM25 as well as two already finetuned DPR retrievers from \cite{karpukhin2020dense, glass-etal-2021-robust}. BM25 is a well established algorithm for retrieval, outperforming until very recently more sophisticated neural-based approaches. It is also particularly efficient and does not require any training, which makes it a very attractive option for real-world production settings. Nevertheless, it is a statistical, pattern matching based approach lacking the ability to learn semantics or context.

Regarding the general domain DPR models, since they are currently state of the art in the relevant general domain tasks, we seek to see if they can be used successfully for the biomedical domain. Our model is trained on far less data, which is nevertheless domain and task specific, therefore it is crucial to understand which approach fares better.

\subsubsection{Experimental Setup}

We employ a PubMed dump from April 2020 as our knowledge sourse, filtering to documents that have an abstract and splitting abstracts to roughly 100-token length passages. We also use a smaller subset of one million passages, in order to be able to search for optimal hyper-parameters and allow easy replication of results. In that subset, we randomly sample passages and add the gold passages from BioSF so as to make sure that a perfect retrieval algorithm would be able to retrieve all correct passages and find the answer. We highlight that this is an easier version of the real-world task, where the retriever needs to search among around 29 million passages.

For BM25, we employ the anserini package \cite{10.1145/3077136.3080721}, and build a Lucene index on the pre-processed passages, whereas we used the off the shelf Huggingface models \cite{} for the general domain DPR retrievers. 

For our retriever, we train DPR on the BioSF dataset, for 40 epochs keeping the best model in terms of the validation loss. We use a learning rate of $3e-5$, an Adam optimizer with default options and a training batch size of 32 examples. Subsequently, we encode the passages with the trained passage encoder. Encoding the full 29 million passages takes around 96 GPU hours on a V100. We then build a flat FAISS \cite{johnson2019billion} index for the encoded passages.

\subsubsection{Results}

Initially, we conduct experiments on the smaller dataset that we described above of one million passages. As we noted in Section \ref{sec:gen_vs_biomedical_domain} evaluating retrieval for slot filling or more broadly for QA in the biomedical domain is significantly different than in the general domain since in biomedicine a query has in most cases multiple answers as opposed to the general domain. Table \ref{tbl:retrieval_on_1m} illustrates the results for this first series of experiments. 

As we can see the DPR models that have been finetuned on the general domain perform rather poorly compared to the much lighter and computationally efficient BM25. Nevertheless, our model Bio-DPR, is substantially better than BM25 in all cases, achieving up to 19 points of improvement (in the case of hits@10). These results, are aligned to the results previously presented for the general domain where BM25 has been outperformed by DPR. Nevertheless, in-domain training data seems critical for DPR to perform well for slot filling, a finding also shared in \cite{maillard2021multi}. 

The same findings apply for the full PubMed knowledge source, as illustrated in Table \ref{tbl:retrieval}, although the improvement of our model over BM25 is much smaller but still significant.

\subsection{Slot Filling Reader}
\label{sec-exp-reader}

For the reader, we finetune a BioBERT-base and a BioBERT-large model on the BioSF training set. We further include two baselines, one trained on the BioASQ 8 QA dataset and one trained in the zero-shot RE (zsRE) dataset from \cite{levy2017zero}. We employ these two baselines to test whether in-domain data from a different task (BioASQ) or general domain data for the same task (zsRE) can be helpful in learning an accurate model.

For all models, we train up to ten epochs, keeping the best performing model on the development set, using a learning rate of $3e-5$, a batch size of $32$ and the Adam optimizer with default parameters. Table \ref{tbl:reader_on_slot} presents the results. We observe that the baselines perform rather poorly compared to the models trained with in-domain slot filling data - a finding that highlights the importance of building an in-domain dataset for slot-filling.

\begin{table*}
\centering
\begin{tabular}{cccc}
\hline 
Model&Data&Exact Match(dev/test)&F1(dev/test) \\ 
\hline
\hline

BioBERT-base&BioASQ&13.10/13.44&17.95/18.64\\
"&zsRE&16.59/15.77&22.51/22.98\\
"&BioSF&52.30/54.67&58.82/59.98\\
\hline 
BioBERT-large&BioSF&\textbf{54.80/55.65}&\textbf{60.92/61.55}\\
\hline 
\end{tabular}
\caption{\label{tbl:reader_on_slot} Evaluation results for the reader experiments on the BioSF development and testing sets. We report the averages across five runs for each model, results in bold show a statistically significant improvement for a z-test at p-value of 0.05.}
\end{table*}

\subsection{End to End Evaluation}
\label{sec:e2e_eval}

Having evaluated both components of our approach, we now turn our attention to the end to end setting, which simulates better a real world scenario. In this setting, we are given a head entity and a relation and we want to correctly extract the tail entity. To evaluate our approach in such a setting, we first use the triples included in the BioSF test set. This dataset contains 3,171 queries with 2.35 answers, i.e. tails, per query on average.

Additionally, we would like to understand how our approach performs in the zero-shot setting, i.e., for entities and relations that our model has not seen during training. To this end, we employ Hetionet \cite{himmelstein2017systematic}, a network of biomedical knowledge assembled from 29 biomedical Knowledge Bases, containing 24 distinct relation types. We keep nine relation types that our models have not previously seen, e.g., "expresses", "localizes", "treats" and randomly sample 500 queries, with 9.3 answers, i.e. tails, per query on average. We note that this dataset differs substantially to the previous one, in the sense that a query might have far more valid answers. For example, some queries have more than 100 valid answers.

\begin{table*}
\centering
\begin{tabular}{cc|cc}
\hline 

Setting&Dataset&end-to-end micro-recall\\
\hline
\hline
Standard&BioSF test set&24.38\\
Zero-shot&Hetionet&18.66\\
\hline 
\end{tabular}
\caption{\label{tbl:e2e_results} End to end evaluation of our approach on a standard as well as a zero-shot setting. }
\end{table*}

In both cases, we first retrieve the top-100 passages for each query, from the full PubMed knowledge source, using our bio-DPR model and subsequently we pass all query-passage pairs through our reader model. We evaluate with micro-recall since, as we discussed previously, there might be multiple valid answers not contained in our KB and we aim to examine what percentage of the KB triples we can extract from text. We note again that this is not a perfect evaluation as, besides the issue mentioned above, there might also be triples in Hetionet that do not appear in any sentence in the literature. Table \ref{tbl:e2e_results} illustrates our results. The recall is substantially low, a finding that is somewhat expected due to the imperfect nature of our evaluation setting, as well the challenging nature of the task, especially in the zero-shot setting. Nevertheless, we consider that these two additional datasets, will enable further research and improved approaches.
Overall, the above experiments should be regarded as a stepping stone towards a novel paradigm for biomedical IE, overcoming the shortcomings of the current standard approach.

\section{Conclusions and Future Work}
\label{sec:conclusions}
In this work we formulated the task of biomedical Information Extraction as a slot filling problem. This approach aims to forgo the need for entity and relation type specific training data, which is scarce and costly to annotate in the biomedical domain. Additionally, this formulation allows to deal with the addition of new relation types, without needing to re-train the relevant models.

Additionally, we have introduced a new biomedical slot filling benchmark and used it to train a biomedical DPR model, a dual BERT-based encoder for retrieval, as well as a biomedical slot filling reader based on BioBERT. In a series of experiments our approach outperforms significantly a number of general domain baselines as well as the simpler BM25 retriever. Furthermore, our results illustrate the importance of in-domain, task-specific training data, in line with findings from recent works \cite{glass-etal-2021-robust, maillard2021multi}.

In future work, we aim to focus on sequence to sequence variants of this work such as the work in \cite{izacard2021leveraging}, as well as to conduct a thorough comparison of a standard biomedical IE system against our slot filling approach.


\bibliographystyle{acl_natbib}
\bibliography{main}
\end{document}